\documentclass{article}

\usepackage{times}  
\usepackage{helvet}  
\usepackage{courier}  
\usepackage{url}  
\usepackage{graphicx}  

\usepackage{amsthm}

\usepackage{url}

\usepackage{amssymb,amsmath,amsthm}

\usepackage{amsfonts}
\usepackage{amssymb}
\usepackage{mathbbol}

\usepackage{mathtools}

\usepackage{multirow}

\usepackage{mdframed}
\usepackage{lipsum}

\usepackage[utf8]{inputenc}
\usepackage[english]{babel}
\usepackage[whole]{bxcjkjatype}

\usepackage{cleveref}

\newtheorem{definition}{Definition}
\newtheorem{theorem}{Theorem}
\newtheorem{lemma}{Lemma}

\newtheorem{assumption}{Assumption}

\crefname{lemma}{lemma}{lemmas}
\Crefname{lemma}{Lemma}{Lemmas}
\crefname{theorem}{theorem}{theorems}
\Crefname{theorem}{Theorem}{Theorem}
\Crefname{definition}{Definition}{Definition}

\begin{document}

%

\title{
Word2Vec is a special case of Kernel Correspondence Analysis and Kernels for Natural Language Processing
}
\author{
  Hirotaka Niitsuma
  \\
  Department of Computer Science\\
  Okayama University\\
  Okayama, Japan\\
  \texttt{niitsuma@cs.okayama-u.ac.jp} \\
  \and
  Minho Lee
  \\
  Graduate School of Science and Technology\\
  Kyungpook National University\\
  Daegu, South Korea\\
  \texttt{mholee@knu.ac.kr}\\
}

\maketitle

\begin{abstract}
We show that correspondence analysis (CA) is equivalent to defining a Gini index with appropriately scaled one-hot encoding.
Using this relation, we introduce a nonlinear kernel extension to CA.
This extended CA gives a known analysis for natural language via specialized kernels that use an appropriate contingency table.
We propose a semi-supervised CA, which is a special case of the kernel extension to CA.
Because CA requires excessive memory if applied to numerous categories, CA has not been used for natural language processing.
We address this problem by introducing delayed evaluation to randomized singular value decomposition.
The memory-efficient CA is then applied to a word-vector representation task.
We propose a tail-cut kernel, which is an extension to the skip-gram within the kernel extension to CA.
Our tail-cut kernel outperforms existing word-vector representation methods.
\end{abstract}

\section{Introduction}\label{sec:introduction}

Principal component analysis (PCA) is a form of unsupervised feature extractor.
When applied to chi-squared distances of categorical data, PCA becomes correspondence analysis (CA).
CA can extract numeric vector features from categorical data without supervised labeling.
The simplest numerical representation for categorical data is a histogram-based representation such as tf-idf or a ``bag-of-words''.
Many applications use such simple representations.
However, histogram-based representations cannot make use of information about correlations within the data.
CA enables the representation of both histograms and correlations in data.

The most popular problem involving categorical data is natural language processing (NLP).
However, CA has not been applied to NLP because most NLP problems involve a large number of categories.
For example, the entire Wikipedia text comprises more than 10,000 different words.
Because CA requires excessive memory if applied to numerous categories, CA has not been used for NLP problems involving more than 10,000 categories.

CA is implemented by singular value decomposition (SVD) of a contingency table.
In many categorical problems, the contingency table is sparsely populated.
Randomized SVD~\cite{Halko:2011:FSR:2078879.2078881} is an appropriate SVD method for sparse matrices.
However, CA requires dense matrix computation even for a sparse contingency table, which makes very large demands on memory resources.
This research aims to address this problem.
We propose using a randomized SVD with delayed evaluation to avoid expanding the sparse matrix into a dense matrix.
We refer to this process as the delayed sparse randomized SVD (DSSVD) algorithm.
We show that CA with DSSVD can be applied to NLP problems.

Neural-network-based approaches are the most popular feature extractors used in NLP.
Of these, \verb|word2vec|~\cite{NIPS2013_5021} is well known.
Usually, such an approach will involve many parameters, which do not have explicit meanings in most cases.
These parameters have to be tuned by grid searching or manual parameter tuning, which is difficult in the absence of explicit meanings for the parameters.
This parameter problem with neural-network-based approaches also gives rise to domain problems.
For example, if \verb|word2vec| is tuned for application to restaurant reviews, the tuning may not be appropriate for movie reviews.

In most cases, the weight values used in neural networks are initialized as random values, which means that the computed results will always be different.
For example, word-vector representations using \verb|word2vec| will always be different, even when the same parameter values are used, because of random initial values.
This adds to the difficulty of parameter tuning.

Since CA is PCA of a contingency table, always the same result is computed.
From this viewpoint, the CA approach is better than neural-network-based approaches.
Although the latter have these issues, they can be used to approximate any nonlinear function, which means that they can be used for a wide variety of problems.
However, because CA is a form of linear analysis, it is not directly applicable to nonlinear problems.
To address this issue, this research introduces a nonlinear kernel extension to CA.
We can then show that this nonlinear CA approach is better in accuracy compared with recent neural-network-based approaches.
In particular, we focus on comparison with respect to word-vector representation tasks.
To distinguish linear and nonlinear CA, we refer to the linear CA as LCA.



\section{CA}

\begin{figure}[ht]
  \centering
  \includegraphics[width=6cm]{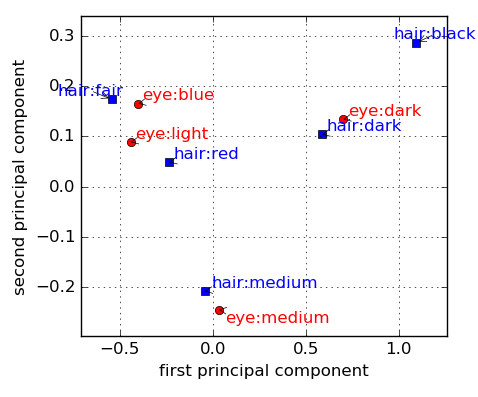}
  \caption{Visualizing Fisher's data.}
  \label{fig:fisherCA}
\end{figure}

CA is a statistical visualization method for picturing the associations between the levels of a two-way contingency table.
As an illustration, consider the contingency table shown in Table \ref{table:Fisher'sdata}. This is well known as ``Fisher's data''~\cite{fishersData1940} and represents the eye and hair color of people in Caithness, Scotland.
The CA of these data yields the graphical display presented in Figure \ref{fig:fisherCA}, 
which shows the correspondence between eye and hair color.

Table \ref{table:Fisher'sdata} shows the joint population distribution of the categorical variable for eye color:
\[
x^{\text{eye}} \in   \{\mbox{ blue light medium  dark} \}.
\]
 and the categorical variable for hair color:
\[
  x^{\text{hair}} \in
\{\mbox{ fair red medium dark black} \}
\]
The visualization is based on ``one-hot encodings'' and the ``indicator matrices'' of categorical variables.
For example, a one-hot encoding ${\bf e}^{\text{eye}}$ and indicator matrix ${H^{\text{eye}}}$ of the categorical variable $x^{\text{eye}}$ can be defined as:
\begin{align}
{\bf e}^{\text{eye}}(\text{blue})=  (1,0,0,0)^t 
\nonumber 
\\
{\bf e}^{\text{eye}}(\text{light}) =  (0,1,0,0)^t 
\nonumber 
\\
{\bf e}^{\text{eye}}(\text{medium})=(0,0,1,0)^t 
\nonumber 
\\
{{\bf e}^{\text{eye}}}(\text{dark})=(0,0,0,1)^t 
\end{align}

\begin{equation}
{H^{\text{eye}}}= 
\begin{bmatrix}
{\bf e}^{\text{eye}}(\text{eye color of 1st person}))^t \\
...\\
0, 0, 1, 0\\
...\\
{\bf e}^{\text{eye}}(\text{eye color of}\;  \text{$n$-th person}))^t
\end{bmatrix}.
\end{equation}

In the following, $x^{\text{r}}$ and $x^{\text{c}}$ denote categorical variables representing the row and column of a contingency table, respectively.
${\bf e}^{\text{r}}(x^{\text{r}}(a))$ and ${\bf e}^{\text{c}}(x^{\text{c}}(a))$ denote one-hot encodings of $x^{\text{r}}$ and $x^{\text{c}}$ for the $a$-th instance.

\begin{table}[ht]
  \caption{Fisher's data.}
  \label{table:Fisher'sdata}
\begin{center}
$x^{\text{hair}}$\\
$x^{\text{eye}}$
\begin{tabular}{cccccc}
 &fair &red &medium &dark &black\\
blue &326& 38& 241& 110& 3\\
light &688& 116& 584& 188& 4\\
medium &343 &84 &909 &412 &26\\
dark &98 &48 &403 &681 &85
\end{tabular}
\end{center}
\end{table}

\section{Covariance Based on Gini Index }

Consider the following relation~\cite{6260326} about the variance of continuous data.

\begin{lemma}
The variance of continuous data can be expressed as the sum of the differences of individual instances:
\begin{align}
\text{\normalfont Var}(x)
&=\frac{1}{n} \sum_{a=1}^n (x(a)-\bar{x})^2
\label{eq:variance-relation-1}
\\
&=\frac{1}{2n^2}\sum_{a=1}^n \sum_{b=1}^n (x(a)-x(b))^2
\label{eq:variance-relation}
\end{align}
where $\{x(1),x(2),...,x(n)\}$ are continuous sample data.
$\bar{x}=\frac{\sum_{b=1}^n x(b)}{n}$ is the average value of the sample data.
$\text{\normalfont Var}(x)$ is the variance of the continuous data.
\end{lemma}
\begin{proof}
Let us expand (\ref{eq:variance-relation-1}) and (\ref{eq:variance-relation}).
\begin{align}
&\frac{1}{n} \sum_{a=1}^n (x(a)-\bar{x})^2
\nonumber\\
&=\frac{1}{n}\left(   \sum_{a=1}^n   x(a)^2 +\bar{x}^2 - 2 x(a) \bar{x}  \right)
\nonumber\\
&=\frac{1}{n} \left(  (\sum_{a=1}^nx(a)^2) + n \bar{x}^2 - 2 n \bar{x} \bar{x}  \right) 
\nonumber\\
&=\frac{1}{n}  \left((\sum_{a=1}^n x(a)^2) - n \bar{x}^2  \right) 
\label{eq:variance-relation-1-expand}
\\
&\frac{1}{2n^2}\sum_{a=1}^n \sum_{b=1}^n (x(a)-x(b))^2
\nonumber\\
&=\frac{1}{2n^2}\sum_{a=1}^n \sum_{b=1}^n \left( (x(a)^2 + x(b)^2 -2 x(a) x(b) )\right) 
\nonumber\\
&=\frac{1}{2n^2}
\left(
2 n \sum_{a=1}^n x(a)^2
- 2 \sum_{a=1}^n x(a)( \sum_{b=1}^n x(b) )
\right) 
\nonumber\\
&=\frac{1}{2n^2}
\left(
2 n \sum_{a=1}^n x(a)^2
- 2 \sum_{a=1}^n x(a) n \bar{x}
\right) 
\nonumber\\
&=\frac{1}{2n^2}
\left(
2 n \sum_{a=1}^n x(a)^2
- 2 n \bar{x} n \bar{x}
\right) 
\nonumber\\
&=\frac{1}{2n^2}
\left(
2 n \sum_{a=1}^n x(a)^2
- 2 n^2 \bar{x}^2
\right) 
\nonumber\\
&=\frac{1}{n}
\left(
\sum_{a=1}^n x(a)^2
-  n \bar{x}^2
\right) 
\label{eq:variance-relation-expand}
\end{align}
We see the expanded the equations (\ref{eq:variance-relation-1-expand}) and (\ref{eq:variance-relation-expand}) are same. 
\end{proof}

Using the same formulation about the sum of the differences between individual instances for categorical data gives a Gini index~\cite{Gini71}:
\begin{equation}
\text{\normalfont Gini}(x)=\frac{1}{2n^2}\sum_{a=1}^n \sum_{b=1}^n |x(a)-x(b)|
\label{eq:gini-def-abs}
\end{equation}
where
\begin{equation}
|x(a)-x(b)|=
\begin{cases}
1 & x(a) \ne x(b)  \\
0  & x(a) = x(b).
\end{cases}
\label{eq:distCategDef}
\end{equation}
Here, $x$ is a categorical variable that takes one of the values in $\{x_1,x_2,...,x_m \}$.
Rewriting this formulation with one-hot encoding gives:
%
\begin{align}
&{\bf e}(x_1) =  (1,0,0,0,...)^t 
\nonumber 
\\
&{\bf e}(x_2)  =  (0,1,0,0,...)^t.
\nonumber
\\
&...
\nonumber \\
&{\bf e}(x_m)  =  (0,0,0,...,1)^t 
\end{align}
This is more similar to the continuous case:
\begin{equation}
\text{\normalfont Gini}(x)=\frac{1}{2n^2}\sum_{a=1}^n \sum_{b=1}^n \frac{|{\bf e}(x(a))-{\bf e}(x(b))|^2}{2}.
\label{eq:gini-def-onehot}
\end{equation}
Using this one-hot encoding, we can also define the covariance of categorical data.
Consider the categorical variables
$x^{\text{r}} \in \{x_1^{\text{r}},x_2^{\text{r}},...,x_{m^{\text{r}}}^{\text{r}} \} $ and
$x^{\text{c}} \in \{x_1^{\text{c}},x_2^{\text{c}},...,x_{m^{\text{c}}}^{\text{c}} \} $.
If the given sample categorical data is
\begin{align}
\{(x^{\text{r}},x^{\text{c}})\}=
&\{(x^{\text{r}}(1),x^{\text{c}}(1)),(x^{\text{r}}(2),x^{\text{c}}(2)),...,
\nonumber 
\\
&(x^{\text{r}}(a),x^{\text{c}}(a)),...,(x^{\text{r}}(n),x^{\text{c}}(n))\},
\end{align}
we can define the covariance of $x^{\text{r}}$ and $x^{\text{c}}$:
\begin{align}
\text{Cov}^{\text{wrong}}(x^{\text{r}},x^{\text{c}})=\frac{1}{4n^2}\sum_{a=1}^n \sum_{b=1}^n |{\bf e}(x^{\text{r}}(a))-{\bf e}(x^{\text{r}}(b))|
\nonumber\\
\cdot  |{\bf e}(x^{\text{c}}(a))-{\bf e}(x^{\text{c}}(b))|.
\label{eq:cov-def-only-onehot}
\end{align}

\begin{table}[tb]
\caption{Contingency table with high correlation~\cite{Okada2000cov,NiitsumaPakddO05}.}
\label{table:contigency-high-correlation}
\begin{center}
\begin{tabular}{|cc|ccc|}
\hline
 &  & \multicolumn{3}{c|}{$x^{\text{\normalfont c}}$} \\
 &  & $x_1^{\text{\normalfont c}}$ & $x_2^{\text{\normalfont c}}$ & $x_3^{\text{\normalfont c}}$ \\
\hline
\multirow{3}{*}{$x^{\text{\normalfont r}}$}
  & $x_1^{\text{\normalfont r}}$ & 100 & 0&  0 \\
  & $x_2^{\text{\normalfont r}}$ & 0 & 100&  0 \\
  & $x_3^{\text{\normalfont r}}$ & 0 & 1&  100 \\
\hline
\end{tabular}
\end{center}
\end{table}

Okada~\cite{Okada2000cov,NiitsumaPakddO05} showed that this definition is invalid by considering the contingency table shown in Table \ref{table:contigency-high-correlation}.
In this contingency table, $x^{\text{r}}$ and $x^{\text{c}}$ are highly correlated and the instance $(x^{\text{r}},x^{\text{c}})=(x^{\text{r}}_3,x^{\text{c}}_2)$ reduces the correlation between $x^{\text{r}}$ and $x^{\text{c}}$.
However, the instance $(x^{\text{r}},x^{\text{c}})=(x^{\text{r}}_3,x^{\text{c}}_2)$ increases the covariance in the formulation (\ref{eq:cov-def-only-onehot}).
To avoid such an invalid increase, Okada defined the covariance using rotated one-hot encoding~\cite{NiitsumaPakddO05}.

\begin{definition}
The covariance of categorical variables  $x^{\text{r}}$ and $x^{\text{c}}$
is the maximized value :
\begin{align}
&
\text{Cov}(x^{\text{r}},x^{\text{c}})
\nonumber 
\\
& =  \underset{R}{\text{maximize}} &
\frac{1}{2n^2} 
\sum^n_{a=1} \sum^n_{b=1}  
&\frac{1}{2}
({\bf e}^{\text{r}}(x^{\text{r}}(a))-{\bf e}^{\text{r}}(x^{\text{r}}(a)))^t 
\nonumber 
\\
&& & R
({\bf e}^{\text{c}}(x^{\text{c}}(a))-{\bf e}^{\text{c}}(x^{\text{c}}(b)) 
\nonumber 
\label{eq:rotated-cov}
\\
& \quad \text{subject to} &
{R}^t R  ={\bf E}
\end{align}

where $R$ is a rotation matrix that maximizes the covariance.
The vectors ${\bf e}^{\text{r}}(x^{\text{r}})$ and ${\bf e}^{\text{c}}(x^{\text{c}})$ are one-hot encodings of $x^{\text{r}}$ and $x^{\text{c}}$.
\end{definition}

In terms of this definition, the instance $(x^{\text{r}},x^{\text{c}})=(x^{\text{r}}_3,x^{\text{c}}_2)$ reduces the covariance.
In this respect, this definition is better than (\ref{eq:cov-def-only-onehot}).
Expanding the maximization problem, (\ref{eq:rotated-cov}) gives a simplified form.

\begin{lemma}
The maximization problem (\ref{eq:rotated-cov}) is equivalent to

\begin{align}
\text{\normalfont Cov}(x^r,x^c)
  =&
  \underset{R}{\text{\normalfont maximize}}     
& 
&\frac{1}{2} 
\text{\normalfont tr}(R^t \Xi )
\nonumber 
\label{eq:rotated-cov2}
\\
   &
     \text{\normalfont  subject to}
  &
&{R}^t R  ={\bf E},
\end{align}
where 
\begin{align}
&{\Xi}={\bf N}/n- {\bf r} {\bf c}^t/n^2,
\label{eq:N-sub-rc}
\\
&{\bf r} = {\bf N} {\bf 1},
\\
&{\bf c} = {\bf N}^t {\bf 1}.
\end{align}
${\bf N}$ is an $m^{\text{\normalfont r}} \times m^{\text{\normalfont c}}$ contingency table, with entries $n_{ij}$ giving the frequency with which row categorical variable $x^{\text{\normalfont r}}=x_i^{\text{\normalfont r}}$ occurs together with column categorical variable $x^{\text{\normalfont c}}=x_j^{\text{\normalfont c}}$.
${\bf r}$
denotes the vector of row marginals and 
${\bf c} $
is the vector of column marginals.
${\bf 1}=(1,1,1...)^t$.
\end{lemma}

\begin{proof}
Consider the expansion of (\ref{eq:rotated-cov}):
\begin{align}
&\frac{1}{2n^2}
\sum^n_{a=1} \sum^n_{b=1}  
\frac{1}{2}
( {\bf e}^r(x^r(a))-{\bf e}^r(x^r(b)))^t 
\nonumber\\  
& \qquad  \qquad \qquad 
R
({\bf e}^c(x^c(a))-{\bf e}^c(x^c(b)))
\nonumber \\
& =
\frac{1}{4n^2}
\text{\normalfont tr} ( R^t
\sum^n_{a=1} \sum^n_{b=1}  
({\bf e}^r(x^r(a))-{\bf e}^r(x^r(b)))
\nonumber\\ & \qquad \qquad \qquad
 ({\bf e}^c(x^c(a))-{\bf e}^c(x^c(b)))^t
)
\nonumber \\ & =
\text{\normalfont tr}(R^t (
\frac{1}{2n}\sum^n_{a=1} {\bf e}^r(x^r(a)){\bf e}^c(x^c(a))^t
\nonumber\\ & \qquad
 -  \frac{1}{2n^2}\sum^n_{a=1} {\bf e}^r(x^r(a)) \sum^n_{b=1}  {\bf e}^c(x^c(b))^t
))
\nonumber \\ & = 
\frac{1}{2}\text{\normalfont tr}(R^t (\frac{{H^r}^t H^c }{n}- \frac{ {\bf r} {\bf c}^t}{n^2})) 
\nonumber \\ & =
\frac{1}{2}\text{\normalfont tr}(R^t (\frac{\bf N}{n}- \frac{ {\bf r} {\bf c}^t}{n^2})) = \frac{1}{2} \text{\normalfont tr}(R^t {\Xi}),
\label{eq:Xi-expand}
\end{align}
where 
\begin{align}
{H^r}=[{\bf e}^r(x^r(a_1)),{\bf e}^r(x^r(a_2)),...,{\bf e}^r(x^r(a_n))]^t
\nonumber\\
{H^c}=[{\bf e}^c(x^c(a_1)),{\bf e}^c(x^c(a_2)),...,{\bf e}^c(x^c(a_n))]^t.
\end{align}
${H^r}$ is the $n \times m^r$ indicator matrix of $x^r$. 
${H^c}$ is the $n \times m^c$ indicator matrix of $x^c$. 
The contingency table ${\bf N}$ can be constructed using the matrix product of the two indicator matrices:
\begin{equation}
{\bf N}={H^r}^t {H^c}.
\end{equation}
Substituting (\ref{eq:Xi-expand}) into
(\ref{eq:rotated-cov})
gives (\ref{eq:rotated-cov2}).
\end{proof}

We can solve the maximization problem (\ref{eq:rotated-cov2}) using SVD.

\begin{theorem}
\label{th:gini-svd}
$R=U {V}^t$
is a local optimum of the maximization problem 
(\ref{eq:rotated-cov2}).
Here,
\begin{equation}
U S {V}^t=  {\Xi}
\label{eq:svd-Xi}
\end{equation}
is an SVD of ${\Xi}$.
\end{theorem}

\begin{proof}
Local optima of the maximization problem are given by differentiating the Lagrangian:

\begin{equation}
{\cal L}=\text{\normalfont tr}(R^t \Xi ) - \text{\normalfont tr}(\Lambda^t( R^tR-{\bf E} )),
\end{equation}

where $\Lambda$ is a Lagrange multiplier.
The differentiation of this Lagrangian with respect to $R$ gives the stationary condition:

\begin{equation}
 R^t \Xi= (\Lambda +\Lambda^t).
\label{eq:ca-stationary-condition}
\end{equation}

This result shows that $R^t \Xi$ must be a symmetric matrix.

Consider the SVD:

\begin{equation}
U S {V}^t=  {\Xi}
\end{equation}

for the case $R=U {V}^t$.
Here, $R^t \Xi$ is the symmetric matrix:
\[R^t \Xi = V U^t U S V^t = V  S V^t \]
and $R$ is the rotation matrix:
\[
R^t R = V U^t U V^t = {\bf E}.
\]
$R=U {V}^t$ satisfies 
the stationary condition (\ref{eq:ca-stationary-condition}) and the constraint of Lemma 2.
We can therefore conclude that $R=U {V}^t$ is a local optimum of the probleam of Lemma 2.
\end{proof}

\begin{theorem}
\label{th:gini-svd-positive}
When 
all singular values of ${\Xi}$ are positive,
$R=U {V}^t$
is the global optimum for the maximization problem 
(\ref{eq:rotated-cov2}).
\end{theorem}

\begin{proof}
Substituting $U S {V}^t={\Xi}$ into (\ref{eq:rotated-cov2}) gives:
\[
\text{\normalfont tr}(R^t \Xi )
=
\text{\normalfont tr}(R^t U S {V}^t )
=
\text{\normalfont tr}({V}^t R^t U S  ).
\]
Note that 
${V}^t R^t U=Q=[q_{ij}]$ is also a rotation matrix.
Consider
\[
\text{\normalfont tr}(Q S ) =  \text{\normalfont tr}(Q D({\bf s}) ) =\sum_i q_{ii} s_i,
\]
where ${\bf s}=(s_1,s_2,...)$ is the vector of the singular values of ${\Xi}$.
Because $Q$ is a rotation matrix, $\forall_i |q_{ii}| \le 1$.
Then,
\begin{equation}
\sum_i q_{ii} s_i \le \sum_i s_i.
\label{eq:upper-limit-qs-pos}
\end{equation}
The case $Q=E$ gives the upper limit:
\[
\sum_i q_{ii} s_i = \sum_i 1 s_i = \sum_i s_i.
\]
\end{proof}

In our experiments, we did not find a case for which ${\Xi}$ had a large negative singular value.
In the following, we assume that $R=U {V}^t$ is the global optimum of the maximization problem (\ref{eq:rotated-cov2}).

If negative singular values appear, we can use the following theorem.

\begin{theorem}
\label{th:gini-svd-pos-neg}
Consider the following optimization problem for a given matrix $\Xi$:

\begin{align*}
&
\underset{R}{\text{\normalfont maximize}}
& 
&
\text{\normalfont tr}(R^t \Xi )
\nonumber 
\\
&\text{\normalfont subject to} &
&{R}^t R  ={\bf E}.
\end{align*}
The global optimum of this optimization problem is:
\[
R=U D(\text{\normalfont sgn}({\bf s}) ) V^t.
\]
Here,
\begin{equation*}
U D({\bf s}) {V}^t=  {\Xi}
\end{equation*}
is an SVD of the matrix ${\Xi}$, where
${\bf s}=(s_1,s_2,...)$  is the vector of the singular values for the matrix ${\Xi}$.
\end{theorem}

\begin{proof}
Consider the case for which some of the singular values are negative.
In such a case, the upper limit (\ref{eq:upper-limit-qs-pos}) becomes:
\begin{equation}
\sum_i q_{ii} s_i \le \sum_i \text{\normalfont sgn}(s_i) s_i = \sum_i |s_i|.
\label{eq:upper-limit-qs-pos-neg}
\end{equation}
The case $Q=D(\text{\normalfont sgn}({\bf s}) )$ gives the upper limit:
\[
\sum_i q_{ii} s_i = \sum_i  \text{\normalfont sgn}(s_i) s_i = \sum_i |s_i|.
\]
\end{proof}

\section{LCA}

First, we introduce generalized singular value decomposition (GSVD).

\begin{definition}
Generalized singular value decomposition (GSVD) of a given matrix $\Xi$ with diagonal weight matrices $D({\bf r})$ and $D({\bf c})$ is the decomposition:
\begin{equation}
{\breve{U}} \hat{S} \breve{V}^t  = \Xi 
\label{eq:gsvd-decomposition}
\end{equation}
where 
\begin{align}
& \breve{U}=D({\bf r})^{1/2} \hat{U},
\\
& \breve{V}=D({\bf c})^{1/2} \hat{V}.
\end{align}
$D({\bf v})$ denotes the diagonal matrix for which diagonal entries are the components of vector ${\bf v}$.
The vectors ${\bf r}$ and ${\bf c}$ are given weight vectors.
$\hat{U}$ and $\hat{V}$ are given by ordinary SVD:
\begin{equation}
\hat{U} \hat{S} { \hat{V} }^t  = \hat{\Xi}
\label{eq:hatXi-svd}
\end{equation}
where
\begin{equation}
 \hat{\Xi} = D({\bf r})^{-1/2} \Xi D({\bf c})^{-1/2}.
\label{eq:def-hat-Xi}
\end{equation}
\end{definition}

Note that this decomposition maintains the perpendicularity of the base vectors in the decomposed space with the weight matrices:
\begin{equation}
{\breve{U}}^t D({\bf r})^{-1} {\breve{U}} = {\breve{V}}^t D({\bf c})^{-1} {\breve{V}}  ={\bf E}.
\end{equation}

Using GSVD, we can define the well-known analysis for categorical data
\footnote{\url{http://forrest.psych.unc.edu/research/vista-frames/pdf/chap11.pdf}}.

\begin{definition}
The liner correspondence analysis (LCA) of a given contingency table ${\bf N}$
is GSVD:
\begin{equation}
{\breve{U}} \hat{S} \breve{V}^t  = \Xi 
\label{eq:ca-gsvd}
\end{equation}
with weight matrices $D({\bf r})$ and $D({\bf c})$.
Here,
\begin{align}
{\Xi}&={\bf N}/n- {\bf r} {\bf c}^t/n^2,
\\
{\bf r}&=\Xi {\bf 1},
\\
{\bf c}&=\Xi^t {\bf 1}. 
\end{align}
\end{definition}

\begin{lemma}
LCA is equivalent to the maximization problem:
\begin{align}
&\underset{\hat{R}}{\text{\normalfont maximize}}& 
&\frac{1}{2}\text{\normalfont tr}({\hat{R}}^t \hat{\Xi}  )
\nonumber 
\\
&\text{\normalfont subject to}&
& {\hat{R}}^t  \hat{R} ={\bf E},
\label{eq:rot-scaled-xi-cov1}
\end{align}
and has the solution:
\begin{align}
 & \hat{R}=\hat{U} \hat{V}^t. 
\end{align}
Here, 
$\hat{U}$ and $\hat{V}$ are given by ordinary SVD:
\begin{equation}
\hat{U} \hat{S} { \hat{V} }^t  = \hat{\Xi}
\end{equation}
where
\begin{equation}
 \hat{\Xi} = D({\bf r})^{-1/2} \Xi D({\bf c})^{-1/2}.
\end{equation}
\end{lemma}

\begin{proof}
LCA is the GSVD of ${\Xi}$.
The GSVD is the SVD of $\hat{\Xi}$.
Applying \Cref{th:gini-svd} to the SVD of $\hat{\Xi}$ gives the required result.
\end{proof}

(\ref{eq:svd-Xi}) and (\ref{eq:hatXi-svd})
enable the SVD $\hat{U} \hat{S} { \hat{V} }^t  = \hat{\Xi}$ to be rewritten as the following maximization problem based on one-hot encoding.

\begin{theorem}
LCA is equivalent to the maximization problem:
\begin{align}
&
\underset{\breve{R}}{\text{\normalfont maximize}}
& 
&\frac{1}{4n^2} 
\sum_{a,b} 
({\hat{\bf e}}^{\text{\normalfont r}}(x^{\text{\normalfont r}}(a))-{\hat{\bf e}}^{\text{\normalfont r}}(x^{\text{\normalfont r}}(b)))^t
\nonumber 
\\
&&& \quad \quad \breve{R}
({\hat{\bf e}}^{\text{\normalfont c}}(x^{\text{\normalfont c}}(a))-{\hat{\bf e}}^{\text{\normalfont c}}(x^{\text{\normalfont c}}(b)))
\nonumber 
\\
&\text{\normalfont subject to} &
  &\breve{R}^t[ {\hat{\bf e}}^{\text{\normalfont r}}(x_1^{\text{\normalfont r}}), {\hat{\bf e}}^{\text{\normalfont r}}(x_2^{\text{\normalfont r}}),... ]^t \breve{R} [ {\hat{\bf e}}^{\text{\normalfont c}}(x_1^{\text{\normalfont c}}), {\hat{\bf e}}^{\text{\normalfont c}}(x_2^{\text{\normalfont c}}),... ]
\nonumber 
\\
  &&& ={\bf E}
\label{eq:rot-f-cov}
\end{align}
where 
\begin{align}
{\hat{\bf e}}^{\text{\normalfont r}}(x^{\text{\normalfont r}})=D({\bf r})^{-1}{\bf e}^{\text{\normalfont r}}(x^{\text{\normalfont r}}),
\\
{\hat{\bf e}}^{\text{\normalfont c}}(x^{\text{\normalfont c}})=D({\bf c})^{-1}{\bf e}^{\text{\normalfont c}}(x^{\text{\normalfont c}})
\end{align}
are scaled one-hot encodings.
\end{theorem}

\begin{proof}

Substitute the following relations into the maximization problem (\ref{eq:rot-scaled-xi-cov1}):
\begin{align}
& 
\text{\normalfont tr}(\hat{R}^t  \hat{\Xi})
=\text{\normalfont tr}( \breve{R}^t D({\bf r})^{-1}  \Xi D({\bf c})^{-1}  )
\\
&
{\breve{R}}^t  D({\bf r})^{-1} \breve{R}  D({\bf c})^{-1} 
=
D({\bf c})^{1/2}
\hat{R}^t \hat{R}
D({\bf c})^{-1/2} ={\bf E}
\nonumber \\
&\Leftrightarrow
\hat{R}^t \hat{R}={\bf E}
\end{align}
where
\begin{align}
& {\breve{R}} = D({\bf r})^{1/2} \hat{R} D({\bf c})^{1/2}.
\end{align}
Substituting these relations in (\ref{eq:rot-scaled-xi-cov1}) gives:
\begin{align}
&\underset{\breve{R}}{\text{\normalfont maximize}}& 
&\frac{1}{2}\text{\normalfont tr}({\breve{R}}^t D({\bf r})^{-1}  \Xi D({\bf c})^{-1}  )
\nonumber 
\\
&\text{\normalfont subject to}&
& {\breve{R}}^t  D({\bf r})^{-1} \breve{R} D({\bf c})^{-1}={\bf E}.
\label{eq:rot-scaled-xi-cov}
\end{align}
Note that:
\begin{align}
&[{\hat{\bf e}}^r(x^r_1),{\hat{\bf e}}^r(x^r_2),...,{\hat{\bf e}}^r(x^r_{n^r})]
\nonumber \\
&=D({\bf r})^{-1} [{\bf e}^r(x^r_1),{\bf e}^r(x^r_2),...,{\bf e}^r(x^r_{n^r})]
\nonumber \\
&=D({\bf r})^{-1} {\bf E}=D({\bf r})^{-1}
\nonumber \\
&[{\hat{\bf e}}^c(x_1^c),{\hat{\bf e}}^c(x_2^c),...,{\hat{\bf e}}^c(x_{n^c}^c)]
\nonumber \\
&=D({\bf c})^{-1} [{\bf e}^c(x^c_1),{\bf e}^c(x^c_2),...,{\bf e}^c(x^c_{n^c})]
\nonumber \\
&=D({\bf c})^{-1} {\bf E} =D({\bf c})^{-1}.
\end{align}
Substituting this relation into problem (\ref{eq:rot-scaled-xi-cov}) yields the optimization problem (\ref{eq:rot-f-cov}).
\end{proof}

This maximization problem defines the rotated Gini index using scaled one-hot encoding.
We can therefore say that LCA is equivalent to defining a Gini index using scaled and rotated one-hot encoding.

\section{Nonlinear extension}

We now consider extending the optimization problem (\ref{eq:rot-f-cov}) using one-hot encoding on a nonlinear mapped space.

\begin{definition}
A nonlinear extension to CA can be expressed as: 
\begin{align}
&\underset{R}{\text{\normalfont maximize}}& 
&\frac{1}{4n^2} 
\sum_{a,b}
(
\Phi^{\text{\normalfont r}}({\hat{\bf e}}^{\text{\normalfont r}}(x^{\text{\normalfont r}}(a)))
\ominus^{\text{\normalfont r}}
\Phi^{\text{\normalfont r}}({\hat{\bf e}}^{\text{\normalfont r}}(x^{\text{\normalfont r}}(b)))
)^t 
\nonumber\\
&&& \quad \quad \cdot R
(
\Phi^{\text{\normalfont c}}({\hat{\bf e}}^{\text{\normalfont c}}(x^{\text{\normalfont c}}(a)))
\ominus^{\text{\normalfont c}}
\Phi^{\text{\normalfont c}}({\hat{\bf e}}^{\text{\normalfont c}}(x^{\text{\normalfont c}}(b)))
)
\nonumber\\
&\text{\normalfont subject to}& 
&R^t[ \Phi^{\text{\normalfont r}}({\hat{\bf e}}^{\text{\normalfont r}}(x_1^{\text{\normalfont r}})), \Phi^{\text{\normalfont r}}({\hat{\bf e}}^{\text{\normalfont r}}(x_2^{\text{\normalfont r}})),... ]^t 
\nonumber\\
&&& R [ \Phi^{\text{\normalfont c}}({\hat{\bf e}}^{\text{\normalfont c}}(x_1^{\text{\normalfont c}})), \Phi^{\text{\normalfont c}}({\hat{\bf e}}^{\text{\normalfont c}}(x_2^{\text{\normalfont c}})),... ] ={\bf E}
\label{eq:non-liniear-gini1}
\end{align}
where $\Phi^{\text{\normalfont r}}, \Phi^{\text{\normalfont c}}$ are nonlinear mappings.
$ \ominus^{\text{\normalfont r}}, \ominus^{\text{\normalfont c}} $ are subtraction operators on the nonlinear mapped spaces.
The summation operator performs cumulative addition on the nonlinear mapped spaces:
\[
\sum x_i= x_1 \oplus x_2 \oplus ... 
\]
where $ \oplus$ is an addition operator on the nonlinear mapped spaces.
We refer to this formulation (\ref{eq:non-liniear-gini1}) as kernel correspondence analysis (KCA).
\end{definition}

To be able to use the kernel trick, we assume the following rules about subtract and add operations.

\begin{assumption}
\begin{align}
&X (Y_1 \ominus^{\text{\normalfont c}} Y_2)^t
=X Y_1^t  \ominus X Y_2^t
\label{eq:rule-ominus-y}
\\
&(X_1 \ominus^{\text{\normalfont r}} X_1) Y^t
=X_1 Y^t \ominus X_2 Y^t
\label{eq:rule-ominus-x}
\\
&X \ominus (Y  \ominus Z) = (X \oplus   Z) \ominus Y 
\label{eq:rule-ominus-ominus}
\\
&(X \ominus  Y) \oplus  Z  = (X \oplus   Z) \ominus Y
\label{eq:rule-ominus-oplus}
\\
&(X \ominus  Y) \ominus Z  =  X \ominus (Z  \oplus  Y)
\label{eq:rule-ominus-ominus2}
\\
&X \oplus  (Y  \ominus Z) = (X \oplus   Y) \ominus Z
\label{eq:rule-oplus-ominus}
\end{align}
\end{assumption}

Because $ \ominus^{\text{\normalfont r}}, \ominus^{\text{\normalfont c}}, \ominus$, and $\oplus$ are nonlinear operators, these relations are not valid in general. 
However, moving left-hand-side operators to the right-hand side in these relations can move 
$\ominus$ outside the expression.
Moving $\ominus$ to the extreme right enables $\ominus$ to require evaluation only once.

When $\oplus=+$, expanding (\ref{eq:non-liniear-gini1}) using these expansion rules gives the following theorem.

\begin{theorem}
If $\oplus=+$ and the rules in Assumption 1 are valid, 
we can introduce kernel matrices $K^{\text{\normalfont r}}$ and $K^{\text{\normalfont c}}$.
Using the kernel matrices, the maximization problem (\ref{eq:non-liniear-gini1}) becomes:
\begin{align}
&
\underset{R}{\text{\normalfont maximize}}
& 
&\frac{1}{2}\text{\normalfont tr}(  R^t K^{\text{\normalfont r}} (\frac{1}{n}{\bf N} \ominus \frac{1}{n^2}{\bf r} {\bf c}^t)  K^{\text{\normalfont c}})
\nonumber 
\\
&\text{\normalfont subject to} &
&R^t K^{\text{\normalfont r}} R K^{\text{\normalfont c}} ={\bf E}.
\label{eq:KCA-formulation-oplus-is-linear}
\end{align}
\end{theorem}

Note that this formulation requires $\ominus$ to be evaluated only once.

Specifying the operators and kernel matrices enables access to various known analyses for categorical data  and NLP.
Table \ref{table:operators} gives the relation between the specifications and known methods.

\begin{table*}
  \caption{Relations between known methods and kernel specializations.}
  \label{table:operators}
  \centering
  \begin{tabular}{llll}
    Name     & $K^{\text{\normalfont r}}$        & $K^{\text{\normalfont c}}$          & $X \ominus Y$     \\ 
    LCA & $D({\bf r})^{-1}$  &  $D({\bf c})^{-1}$    & $X -Y$         \\
    Gini index ~\cite{Okada2000cov,NiitsumaPakddO05} & ${\bf E}$         &  ${\bf E}$          & $ X-Y $ \\
    SGNS~\cite{NIPS2013_5021,LevyGnips14}  & ${\bf E}$         &  ${\bf E}$          & $  (\log X-\log Y - \log k)$ \\
    GloVe~\cite{pennington2014glove}     & ${\bf E}$         &  ${\bf E}$          & $  (\log X-\log Y + b^w +b^{\text{\normalfont c}})$ \\
  \end{tabular}
\end{table*}

\begin{table*}
\caption{Comparisons using the text8 corpus.}
\label{table:ws-test-text8}

\begin{center}
\begin{tabular}{ll|cccccc} 
& method 
      &  Sim & Rel &  MEN        & M.Turk  & Rare   & S999 \\
  \hline
  & CBOW         &           {0.388} &          {0.438} &          {0.383} &          {0.579} &          {0.050} &          {0.075}  \\
  & SGNS   &           {0.674} &          {0.654} &          {0.561} &          {0.608} &          {0.027} &          {0.215}  \\
  & GloVe  &           {0.431} &          {0.466} &          {0.421} &          {0.508} &          {0.118} &          {0.096}  \\
 &fastText &           {0.655} &          {0.609} &      { 0.636} &          {0.623} &          {0.059} & \textcolor{red}{0.223}  \\
  \hline
  & tail-cut
      & \textcolor{red}{0.762} & \textcolor{magenta}{0.667} & \textcolor{red}{0.682} & \textcolor{magenta}{0.649} &{0.121} &  {0.212} \\

  & LCA  ${\bf N}^{\text{flat}}$
      &\textcolor{magenta}{0.749} & \textcolor{red}{0.680} &  {0.671} & \textcolor{red}{0.668} & \textcolor{magenta}{0.127} & \textcolor{magenta}{0.218} \\ 
  & LCA ${\bf N}^{\text{skip}}$  &           {0.741} &          {0.657} &         \textcolor{magenta}{0.672} &          {0.640} &          \textcolor{red}{0.135} &          {0.211}   \\ 
  \hline
  &SCA+MEN &           {0.743} &          {0.665} &          {0.770} &          {0.636} &          {0.136} &          {0.210} \\ 
  

  & SCA+M.Turk &           {0.741} &          {0.658} &          {0.672} &          {0.798} &          {0.136} &          {0.211} \\ 


 \end{tabular}
\end{center}

\hspace{2cm}
\textcolor{red}{red}: best result
\hspace{2cm}
\textcolor{magenta}{magenta}: 2nd best

\end{table*} 

\subsection{Semi-supervised CA}

Consider the case where we wish to manually tune the distance between one-hot encodings using tuning ratio tables $\gamma^{\text{\normalfont r}}(x^{\text{\normalfont r}},{x^{\prime}}^{\text{\normalfont r}})$ and $\gamma^{\text{\normalfont c}}(x^{\text{\normalfont c}},{x^{\prime}}^{\text{\normalfont c}})$.
\begin{align}
\Phi^{\text{\normalfont r}}({\hat{\bf e}}^{\text{\normalfont r}}(x_a^{\text{\normalfont r}}))
\ominus^{\text{\normalfont r}}
\Phi^{\text{\normalfont r}}({\hat{\bf e}}^{\text{\normalfont r}}(x_b^{\text{\normalfont r}}))
=( {\hat{\bf e}}^{\text{\normalfont r}}(x_a^{\text{\normalfont r}}) - {\hat{\bf e}}^{\text{\normalfont r}}(x_b^{\text{\normalfont r}}) )\gamma^{\text{\normalfont r}}(x_a^{\text{\normalfont r}},x_b^{\text{\normalfont r}})
\nonumber\\
\Phi^{\text{\normalfont c}}({\hat{\bf e}}^{\text{\normalfont c}}(x_a^{\text{\normalfont c}}))
\ominus^{\text{\normalfont c}}
\Phi^{\text{\normalfont c}}({\hat{\bf e}}^{\text{\normalfont c}}(x_b^{\text{\normalfont c}}))
=( {\hat{\bf e}}^{\text{\normalfont c}}(x_a^{\text{\normalfont c}}) - {\hat{\bf e}}^{\text{\normalfont c}}(x_b^{\text{\normalfont c}}) )\gamma^{\text{\normalfont c}}(x_a^{\text{\normalfont c}},x_b^{\text{\normalfont c}})
\label{eq:def-gamma}
\end{align}
For this case, we can define the following problem.

\begin{definition}
Semi-supervised correspondence analysis (SCA) can be expressed as:
\begin{align}
&
\underset{R}{\text{\normalfont maximize}}
& 
&\frac{1}{2n^2}\text{\normalfont tr}(
R^t D({\bf r})^{-1}
(
{\bf N} \circ (\Gamma^{\text{\normalfont r}} {\bf N} \Gamma^{\text{\normalfont c}})
\nonumber 
\\
&&& 
\quad - (\Gamma^{\text{\normalfont r}} {\bf N})  \circ ({\bf N} \Gamma^{\text{\normalfont c}} )
)
D({\bf c})^{-1} )
\nonumber 
\\
&\text{\normalfont subject to} &
&R^t D({\bf r})^{-1} R D({\bf c})^{-1} ={\bf E}
\nonumber 
\\
&&&r=(\Gamma^{\text{\normalfont r}} {\bf N})  \circ ({\bf N} \Gamma^{\text{\normalfont c}} ) {\bf 1}
\nonumber 
\\
&&&c= ( (\Gamma^{\text{\normalfont r}} {\bf N})  \circ ({\bf N} \Gamma^{\text{\normalfont c}} ))^t {\bf 1}
\nonumber 
\\
&&& \Gamma^{\text{\normalfont r}}=[\gamma_{ij}^{\text{\normalfont r}}]
\nonumber 
\\
&&& \Gamma^{\text{\normalfont c}}=[\gamma_{ij}^{\text{\normalfont c}}]
\end{align}
where $\circ$ is the Hadamard product.
\end{definition}

This problem is defined by considering $\oplus=+$ 
and (\ref{eq:def-gamma}). 
The tuning tables  $\gamma^{\text{\normalfont r}}$ and $\gamma^{\text{\normalfont c}}$ can be regarded as supervised training data.
However, this method can also be based on unsupervised training data like PCA.
We refer to this process as semi-supervised correspondence analysis (SCA). 

\section{Delayed Sparse Matrix}

CA is ordinary SVD:
\begin{equation}
\hat{U} \hat{S} { \hat{V} }^t 
=  D({\bf r})^{-1/2} \left( {\bf N}/n- {\bf r} {\bf c}^t/n^2 \right) D({\bf c})^{-1/2}.
\label{eq:hatXi-svd2}
\end{equation}
Because ${\bf r} {\bf c}^t$ is a dense matrix, ${\bf N}/n- {\bf r} {\bf c}^t/n^2$ is also a dense matrix, even when ${\bf N}$ is a sparse matrix.
This is the reason why the CA approach makes such a demand on memory resources.

However, computing the dense matrix can be avoided by delayed evaluation.
Consider multiplying by an arbitrary matrix $Z$ on both the left-hand and right-hand side of (\ref{eq:hatXi-svd2}).
\begin{align}
 &\text{left-dot}(Z) = \text{lamda}(Z)(
\nonumber \\      
  &  \left( Z D({\bf r})^{-1/2} {\bf N}/n- Z D({\bf r})^{-1/2}{\bf r} {\bf c}^t/n^2 \right) D({\bf c})^{-1/2}
    )
\end{align}
The right-hand-side multiplication can be expressed similarly:
\begin{align}
 &\text{right-dot}(Z) = \text{lamda}(Z)(
\nonumber \\      
  &D({\bf r})^{-1/2} \left( {\bf N}/n D({\bf c})^{-1/2} Z - {\bf r} {\bf c}^t/n^2 D({\bf c})^{-1/2} Z \right)
    ).
\end{align}
%
Randomized SVD requires only a multiplying operation on the matrix to be decomposed, as for the power method.
We can execute the randomized SVD using $\text{left-dot}(Z)$ and $\text{right-dot}(Z)$ without involving the expanded matrix (\ref{eq:hatXi-svd2}).
Because this scheme can avoid computing the dense matrix, there is a reduction in both computing time and memory requirements.
We refer to this scheme as the delayed sparse randomized SVD (DSSVD) algorithm.
Python implementation of this CA is provided in \url{https://github.com/niitsuma/delayedsparse/blob/master/delayedsparse/ca.py}

\section{Word Representation}

This research discusses the application of CA to word-vector representation tasks.
Consider the following contingency table for some given training-text data:
\begin{equation}
{\bf N}^{\text{skip}}=[n_{ij}^\text{skip}]=[ \mbox{\#} (w_i,t_j) ],
\label{eq:N-word-context-skip}
\end{equation}
where $\mbox{\#}(w,t)$ is the number of times that the word $w$ appears in the context $t$.
Based on this table, Mikolov et al.~\cite{NIPS2013_5021} introduced vector representations of words, referred to as \verb|word2vec|.
$\mbox{\#}(w,t)$ is computed by using the skip-gram model.
However, the skip-gram model requires random sampling, which gives different results for each computation.
This research uses the following fixed representation.
\[
 \text{\#}(w_1 *^k w_2 )
\]
This notation represents the number of subsentences for which an arbitrary $k$ words appear between words $w_1$ and $w_2$.
For example, consider the sentence:

\; ``this is this is this is this is this.'' \; 

\noindent
In this sentence, the number of times ``is'' occurs three words after ``this'' is:
\[
  \text{\#}(\text{this} * *\; \text{is}) = \text{\#}(\text{this} *^2 \text{is}) = 3.
\]
``is'' also appears in other locations.
\[
\mbox{\#}(\mbox{this} *^0 \mbox{is}) = 4,\;
\mbox{\#}(\mbox{this} *^1 \mbox{is}) = 0,\;
\mbox{\#}(\mbox{this} *^2 \mbox{is}) = 3,
...
\]
Given an appropriate window size $W$, this equation can represent a relation similar to the skip-gram.
\begin{equation}
{\bf N}^{\text{flat}}=[n_{ij}^\text{flat}]=[ \sum_{k=0}^W \text{\#}(w_i *^k w_j )]
\label{eq:N-word-context-flat}
\end{equation}
This  cannot ignore noise relations when $W$ is large.
To ignore noise, we introduce the following weighted sum:
\begin{equation}
{\bf N}^{\text{cut}}=[n_{ij}^\text{cut}]=[ \sum_{k=0}^W \text{\#}(w_i *^k w_j ) \gamma(w_i,w_j,k)]
\label{eq:N-word-context-cut}
\end{equation}
where
\[
\gamma(w_i,w_j,k)=
\begin{cases}
1 \quad  (\text{\#}(w_i *^k w_j ) > \frac{\text{\#}(w_i) \text{\#}(w_j)}{n^2} )  \\
0 \quad  (\text{\#}(w_i *^k w_j ) \le \frac{\text{\#}(w_i) \text{\#}(w_j)}{n^2} ).  \\
\end{cases}
\]
$\text{\#}(w)$ is the number of times that the word $w$ appears in all the training text.
$n$ is the total number of words in the given training text.
The weighted sum can be introduced using a kernel extension similar to SCA.
We refer to this extension as the ``tail-cut kernel''.
We can compute the LCA of ${\bf N}^{\text{skip}}$ and ${\bf N}^{\text{flat}}$
and the KCA of ${\bf N}^{\text{cut}}$.

\section{Experiments}

This section compares various word-vector representation tasks using 
the text8 corpus\footnote{
\url{http://mattmahoney.net/dc/text8.zip}
}
.

\begin{figure}
  \centering
  \includegraphics[width=8.7cm]{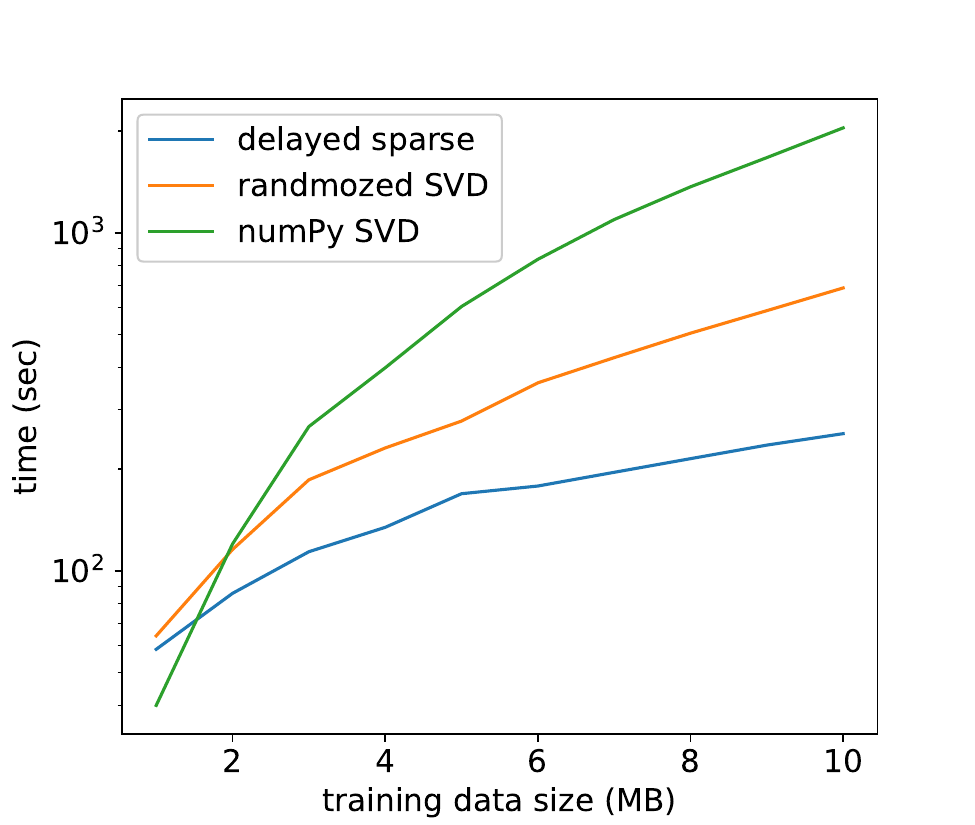}
  
  \caption{SVD comparison time.}
  \label{fig:fsvdcomparetime}
  \centering
  \includegraphics[width=8.7cm]{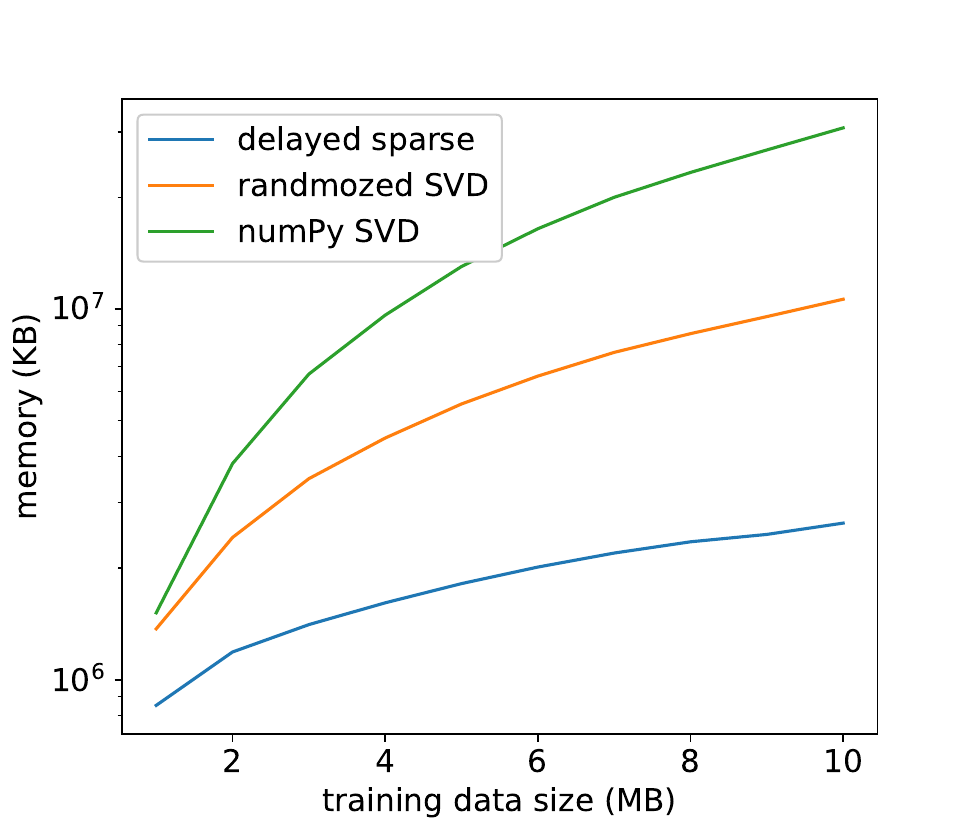}
  \caption{SVD comparison memory.}
  \label{fig:fsvdcomparememory}
\end{figure}

\subsection{Delayed Sparse Randomized SVD}

Figures \ref{fig:fsvdcomparetime} and \ref{fig:fsvdcomparememory} show the computing times and the required memory for LCA, respectively.
The horizontal axis is the size of the training data.
The initial section of the text8 corpus was used as the training data for the LCA.
The experiments were carried out in a Gentoo Linux environment using an Intel i7-3770K 3.50 GHz processor.
Note that the vertical axes have logarithmic scales.

The LCA was computed using SVD with the numPy library, randomized SVD with the the scikit-learn library, and the DSSVD.
DSSVD was 100 times faster than SVD with numPy and 10 times faster than randomized SVD. 
The memory required for DSSVD was 10\% of that required for SVD of numPy and 20\% of that required for randomized SVD.
When using the whole text8 corpus, the differences became more emphatic.
Because of excessive memory requirements, using CA for NLP is impossible without DSSVD.
Python code of this experimets is provided in \url{https://github.com/niitsuma/delayedsparse/blob/master/demo-ca.sh}

\subsection{Word Representation}

We evaluated the English word-vector representation by focusing on the similarity between words using six test datasets.

\begin{description}
\item[Sim:] WordSim Similarity~\cite{Zesch:2008:UWC:1620163.1620206}
\item[Rel:] WordSim Relatedness~\cite{Agirre:2009:SSR:1620754.1620758}
\item[MEN:] MEN dataset~\cite{bruni-EtAl:2012:ACL2012}
\item[M.Turk:]  Mechanical Turk dataset~\cite{Radinsky:2011:WTC:1963405.1963455}
\item[Rare:]  Words dataset~\cite{luong-socher-manning:2013:CoNLL-2013}
\item[S999:] SimLex-999 dataset~ \cite{Hill:2015:SES:2893320.2893324}.
\end{description}

Table \ref{table:ws-test-text8} shows a comparison between methods for the whole text8 corpus.
Evaluation with these six test datasets provided a ranking of similarity among words.
The evaluation values are Spearman's rank correlation coefficients of the ranking of similarity among words.
For comparison, we show the results for skip-gram with negative sampling (SGNS)~\cite{NIPS2013_5021}, 
continuous bag-of-words (CBOW)~\cite{NIPS2013_5021},
GloVe~\cite{pennington2014glove},
and fastText~\cite{bojanowski2016enriching}.

In most cases, the tail-cut kernel provided the best or almost-best results.
The LCA with ${\bf N}^{\text{flat}}$ also provided some of the best results.
However, the LCA with ${\bf N}^{\text{flat}}$ results were drastically affected by the window-size parameter.
LCA for ${\bf N}^{\text{skip}}$ also showed instability, whereas
the tail-cut kernel provided stable results.
For window sizes larger than 30, its result changes become insignificant.
This implies that the tail-cut kernel is relatively independent of the window size parameter, thereby possibly decreasing the number of parameters by one.

SCA based on LCA for ${\bf N}^{\text{skip}}$ were also evaluated.
The SCA used MEN data and M.Turk data as the supervised training data.
SCA outperformed LCA for much of the test data.
These results demonstrate that SCA can work effectively.
Although the word-vector representation task is unsupervised learning,
SCA can use supervised data within the word-vector representation task.
Part of codes of this experiments is provided in \url{https://github.com/niitsuma/wordca}

\section{Conclusion}

We have proposed a memory-efficient CA method based on randomized SVD.
The algorithm also drastically reduces the computation time.
This efficient CA can be applied to the word-vector representation task.
The experimental results show that CA can outperform existing methods in the word-vector representation task.
We have further proposed the tail-cut kernel, which is an extension of the skip-gram approach within KCA.
Again, the tail-cut kernel outperformed existing word-vector representation methods.

\bibliographystyle{plain}
\bibliography{myref.bib}

\begin{thebibliography}{10}

\bibitem{Agirre:2009:SSR:1620754.1620758}
Eneko Agirre, Enrique Alfonseca, Keith Hall, Jana Kravalova, Marius Pa\c{s}ca,
  and Aitor Soroa.
\newblock A study on similarity and relatedness using distributional and
  wordnet-based approaches.
\newblock In {\em Proceedings of Human Language Technologies: The 2009 Annual
  Conference of the North American Chapter of the Association for Computational
  Linguistics}, pages 19--27. Association for Computational Linguistics, 2009.

\bibitem{bojanowski2016enriching}
Piotr Bojanowski, Edouard Grave, Armand Joulin, and Tomas Mikolov.
\newblock Enriching word vectors with subword information.
\newblock {\em arXiv preprint arXiv:1607.04606}, 2016.

\bibitem{bruni-EtAl:2012:ACL2012}
Elia Bruni, Gemma Boleda, Marco Baroni, and Nam~Khanh Tran.
\newblock Distributional semantics in technicolor.
\newblock In {\em Proceedings of the 50th Annual Meeting of the Association for
  Computational Linguistics}, pages 136--145. Association for Computational
  Linguistics, July 2012.

\bibitem{fishersData1940}
R.~A. Fisher.
\newblock The precision of discriminant functions.
\newblock {\em Annals of Eugenics}, 10:422--429, 1940.

\bibitem{Gini71}
C.W. Gini.
\newblock Variability and mutability, contribution to the study of statistical
  distributions and relations. studi economico-giuridici della r. universita de
  cagliari (1912). reviewed in: Light, r.j., margolin, b.h.: An analysis of
  variance for categorical data.
\newblock {\em J. American Statistical Association}, 66:534--544, 1971.

\bibitem{Halko:2011:FSR:2078879.2078881}
N.~Halko, P.~G. Martinsson, and J.~A. Tropp.
\newblock Finding structure with randomness: Probabilistic algorithms for
  constructing approximate matrix decompositions.
\newblock {\em SIAM Review}, 53(2):217--288, May 2011.

\bibitem{Hill:2015:SES:2893320.2893324}
Felix Hill, Roi Reichart, and Anna Korhonen.
\newblock Simlex-999: Evaluating semantic models with genuine similarity
  estimation.
\newblock {\em Comput. Linguist.}, 41(4):665--695, December 2015.

\bibitem{LevyGnips14}
Omer Levy and Yoav Goldberg.
\newblock Neural word embedding as implicit matrix factorization.
\newblock In {\em Proceedings of the 27th International Conference on Neural
  Information Processing Systems}, pages 2177--2185, 2014.

\bibitem{luong-socher-manning:2013:CoNLL-2013}
Thang Luong, Richard Socher, and Christopher Manning.
\newblock Better word representations with recursive neural networks for
  morphology.
\newblock In {\em Proceedings of the Seventeenth Conference on Computational
  Natural Language Learning}, pages 104--113, August 2013.

\bibitem{NIPS2013_5021}
Tomas Mikolov, Ilya Sutskever, Kai Chen, Greg~S Corrado, and Jeff Dean.
\newblock Distributed representations of words and phrases and their
  compositionality.
\newblock In {\em Proceedings of the 26th International Conference on Neural
  Information Processing Systems}, pages 3111--3119. 2013.

\bibitem{NiitsumaPakddO05}
Hirotaka Niitsuma and Takashi Okada.
\newblock Covariance and {PCA} for categorical variables.
\newblock In {\em Proceedings of the 9th Pacific-Asia Conference on Knowledge
  Discovery and Data Mining}, pages 523--528, 2005.

\bibitem{Okada2000cov}
T.~Okada.
\newblock A note on covariances for categorical data.
\newblock In K.S. Leung, L.W. Chan, and H.~Meng, editors, {\em Intelligent Data
  Engineering and Automated Learning - IDEAL 2000}, 2000.

\bibitem{pennington2014glove}
Jeffrey Pennington, Richard Socher, and Christopher~D. Manning.
\newblock Glove: Global vectors for word representation.
\newblock In {\em Proceedings of the 2014 Conference on Empirical Methods in
  Natural Language Processing}, pages 1532--1543, 2014.

\bibitem{Radinsky:2011:WTC:1963405.1963455}
Kira Radinsky, Eugene Agichtein, Evgeniy Gabrilovich, and Shaul Markovitch.
\newblock A word at a time: Computing word relatedness using temporal semantic
  analysis.
\newblock In {\em Proceedings of the 20th International Conference on World
  Wide Web}, pages 337--346, 2011.

\bibitem{Zesch:2008:UWC:1620163.1620206}
Torsten Zesch, Christof M\"{u}ller, and Iryna Gurevych.
\newblock Using wiktionary for computing semantic relatedness.
\newblock In {\em Proceedings of the 23rd National Conference on Artificial
  Intelligence}, pages 861--866, 2008.

\bibitem{6260326}
Y.~Zhang, H.~Wu, and L.~Cheng.
\newblock Some new deformation formulas about variance and covariance.
\newblock In {\em Proceedings of International Conference on Modelling,
  Identification and Control}, pages 987--992, June 2012.

\end{thebibliography}

\end{document}